\documentclass[letterpaper]{article} %
\usepackage[]{aaai25}  %
\usepackage{times}  %
\usepackage{helvet}  %
\usepackage{courier}  %
\usepackage[hyphens]{url}  %
\usepackage{graphicx} %
\usepackage{natbib}  %
\usepackage{caption} %
\usepackage{algorithm}
\usepackage{algorithmic}

\usepackage{newfloat}
\usepackage{listings}
\DeclareCaptionStyle{ruled}{labelfont=normalfont,labelsep=colon,strut=off} %
\floatstyle{ruled}
\newfloat{listing}{tb}{lst}{}
\floatname{listing}{Listing}
\usepackage[textsize=tiny]{todonotes}

\usepackage{booktabs}
\usepackage{tabularx}
\usepackage{multirow}
\usepackage{enumitem}
\usepackage{anyfontsize}
\usepackage{url}
\usepackage{subcaption}
\usepackage{xr}
\usepackage{bbold}
\usepackage{MnSymbol,wasysym}
\usepackage{fontawesome}

\usepackage{times}
\usepackage{latexsym}
\usepackage{xspace}

\usepackage{graphicx}
\usepackage{xfp}

\usepackage{array}
\usepackage{tcolorbox}
\usepackage{makecell}
\usepackage{soul}
\usepackage{stackengine}
\newcolumntype{H}{>{\setbox0=\hbox\bgroup}c<{\egroup}@{}}

\newcommand{\vtwo}[1]{#1}
\newcommand{\vthree}[1]{#1}

\newcommand{\Skip}[1]{}

\newcommand{\modelname}{\textsc{Mera}\xspace}

\newcommand{\ie}{\textit{i}.\textit{e}.\ }
\newcommand{\eg}{\textit{e}.\textit{g}.\ }

\newcommand{\secref}[1]{\S\ref{#1}}
\newcommand{\appref}[1]{Appendix~\ref{#1}}

\newcommand{\tbref}[1]{Table~\ref{#1}}
\newcommand{\equationref}[1]{Equation~\ref{#1}}
\newcommand{\equationrefnoprefix}[1]{\ref{#1}}

\newcommand{\dotieconcat}[2]{%
  \text{\raisebox{.8ex}{$\smallfrown$}}%
}
\newcommand{\mypar}[1]{\paragraph{#1}}

\newcommand{\spacemagic}[1]{}

\newcommand{\reftoapp}[1]{}

\newcommand{\colorcelldneg}[3]{-#1}
\newcommand{\pd}[1]{
\colorcelldneg{#1}{0}{17.5}
}

\title{Memorize and Rank: Elevating Large Language Models for Clinical Diagnosis Prediction}

\author{
    Mingyu Derek Ma$^{1}$, 
Xiaoxuan Wang$^{1}$, 
Yijia Xiao$^1$, 
Anthony Cuturrufo$^1$  \\
\textbf{Vijay S Nori}$^2$, 
\textbf{Eran Halperin}$^1,2$,
\textbf{Wei Wang}$^1$ 
\\
}
\affiliations{
    $^1$University of California, Los Angeles \\ $^2$Optum AI \\
    \{ma, xw27, yijia.xiao, acc, weiwang\}@cs.ucla.edu \\
    vijay.nori@optum.com, eran.halperin@uhg.com
}

\usepackage{bibentry}

\begin{document}

\maketitle

\begin{abstract}
    Clinical diagnosis prediction models, when provided with a patient's medical history, aim to detect potential diseases early, facilitating timely intervention and improving prognostic outcomes. However, the inherent scarcity of patient data and large disease candidate space often pose challenges in developing satisfactory models for this intricate task. The exploration of leveraging Large Language Models (LLMs) for encapsulating clinical decision processes has been limited.
We introduce \modelname, a clinical diagnosis prediction model that bridges pertaining natural language knowledge with medical practice. We apply hierarchical contrastive learning on a disease candidate ranking list to alleviate the large decision space issue. With concept memorization through fine-tuning, we bridge the natural language clinical knowledge with medical codes.
Experimental results on MIMIC-III and IV datasets show that \modelname achieves the state-of-the-art diagnosis prediction performance and dramatically elevates the diagnosis prediction capabilities of generative LMs. 

\end{abstract}

\section{Introduction}
\label{sec:intro}

Electronic Health Records (EHR), containing patient status and diagnoses, 
embody valuable domain expertise and clinical operation patterns~\cite{maccrobat}. Clinicians make diagnosis judgments based on their extensive medical knowledge, acquired through years of education from textbooks and literature, as well as their accumulated experience derived from clinical practice. 
Clinical diagnosis prediction aims to predict patients' diseases that are highly likely to be diagnosed in the upcoming hospital admission by analyzing the patients' past diagnoses. The input and output are both presented in sequences of medical codes, which do not directly convey semantic information nor disease property.
The resulting AI-enhanced diagnosis system~\cite{Morid2023TimeSeriesPrediction} may enable 
early warning of diseases~\cite{rochefort2015accuracy}, optimized clinical resource allocation~\cite{yadav2013automated}, and better risk estimation for sustainable insurance~\cite{hsu2016data}.

Two primary challenges in diagnosis prediction have driven various research efforts~\cite{Wornow2023ShakyFoundationsLarge} but remain unsolved. 
First, what would be the best practice to incorporate clinical knowledge into the model? Existing works initialize concept embeddings from natural language descriptions~\cite{Wu2023MEGACareKnowledgeguidedMultiview,Bornet2023ComparingNeuralLanguage}, or enrich patient representation with external disease ontologies~\cite{An2023KAMPNetMultisourceMedical,Cheong2023AdaptiveIntegrationCategorical}. However, a significant gap persists between the primary knowledge modality, \ie natural language, and the model's hidden representation. 
Second, how can we handle the large candidate space when making predictions and exploit the supervisory signals induced from this candidate space? The commonly used International Classification of Diseases (ICD) coding system encodes 13k+ diseases~\cite{Cartwright2013ICD9CMICD10CMCodes}. Existing works typically treat the task as $k$-way classification where $k$ is the number of possible diseases, and then apply cross entropy loss for each disease individually. These approaches often overlook the dependencies among diseases and the structural nuances within \vtwo{the diagnosis coding system}.

Generative Language Models (LM), especially the Large Language Models (LLM), are trained to predict the next token, adhere to task instructions~\cite{Brown2020LanguageModelsAre,Ma_Wang_Kung_Brantingham_Peng_Wang_2024}, and align with human preferences~\cite{Ouyang2022TrainingLanguageModels}. These models exhibit superior capabilities in language understanding and reasoning, as shown by their performance on science-based benchmarks~\cite{clibench,Wu2023PMCLLaMABuildingOpensource,zhang2024climbbenchmarkclinicalbias}. During the pretraining stage, LLMs assimilate a large amount of knowledge extracted from literature and online corpora. However, there remains an underexplored domain in using LLM for clinical diagnosis prediction, due to the aforementioned gap between natural language and medical code, as well as the disparity between the token-level optimization process and the large candidate outcome space. These challenges impede the effective application of generative LMs to diagnosis prediction tasks, even as the state-of-the-art models predominantly rely on graph neural networks without fully harnessing natural language knowledge~\cite{Yang2023InterpretableDiseasePrediction,Wu2023MEGACareKnowledgeguidedMultiview,An2023KAMPNetMultisourceMedical}.
\vthree{Fine-tuning generative LM LLaMA2~\cite{Touvron2023LlamaOpenFoundation} directly on diagnosis prediction yields almost 20-point lower recall@20 than GNN-based existing best model~\cite{Yang2023InterpretableDiseasePrediction} as shown in \tbref{table:diagnosis_pred_main}.}
There are some studies that use transformer-based LM for clinical outcome prediction, but they either do not support structured data as input~\cite{Niu2024EHRKnowGenKnowledgeenhancedMultimodal,Wang2023DRGLLaMATuningLLaMA}, not compatible with mainstream LLMs~\cite{Rupp2023ExBEHRTExtendedTransformer,Guo2023EHRFoundationModels}, or only work for narrow output space with few classes~\cite{Wang2023DRGLLaMATuningLLaMA,Shoham2023CPLLMClinicalPrediction}.

To tackle these challenges, we propose \modelname, an LLM designed for clinical diagnosis prediction that incorporates a comprehensive understanding of clinical knowledge by leveraging relationships among medical codes and offers extensive supervision over the output space. The patient's historical diagnosis results are formulated as linear sequences and the LLM is tasked with generating a probability distribution for the diagnosis results in the subsequent visit. 
Compared with the ordinary paradigm that optimizes the probability of generating the correct token, we optimize the outcome directly.
To enhance the inter-visit causal reasoning, we employ contrastive learning to compel the model to distinguish true diagnoses from false ones. The contrastive learning process is extended to multiple levels in the hierarchical organization of the medical codes within the ICD coding ontology. The model is learned to distinguish the true diagnoses from a pool of potential diagnoses while the pool is increasingly relevant to the true ones. To regularize the diagnosis predictions to follow intra-visit diagnosis patterns, we develop a teaching-forcing strategy to optimize the medical code ranking, assuming partial diagnoses of the visit are known. 
\vtwo{
To allow the model to grasp the comprehensive clinical semantics and diagnosis property of each medical code, we fine-tune the LM to ``memorize'' the mapping between medical codes and their natural language definitions. Consequently, this process bridges the gap between raw codes and their contextual medical meanings and equips the LM to capture the intricate code dependencies that are crucial for precise diagnosis assessments.
}

We validate the effectiveness of \modelname in general diagnosis and heart failure prediction tasks on the patient records in MIMIC-III~\cite{Johnson2016MIMICIIIFreelyAccessible} and MIMIC-IV~\cite{Johnson2023MIMICIVFreelyAccessible} datasets. \modelname yields significant improvements over the existing state-of-the-art models across tasks on all datasets while having almost perfect memorization of bidirectional medical code-definition mapping. An extensive analysis of leading LLM's medical code understanding and diagnosis prediction capabilities is conducted, and we observe that GPT-4 is still far behind fine-tuned models on both tasks.
We further conduct ablation studies to validate the effectiveness of the proposed novel design choices.

\spacemagic{\vspace{-1em}}
\section{Preliminaries}
\spacemagic{\vspace{-0.6em}}

\subsection{Task Formulations}
\label{sec:task_formulations}
\spacemagic{\vspace{-0.6em}}

\vthree{
\modelname can be applied for any task whose output is a collection of candidates belonging to a pre-defined decision space. We introduce widely used diagnosis prediction settings as typical testbeds for \modelname~\cite{Yang2023InterpretableDiseasePrediction}.

\mypar{Tasks.}
The first task is a general \textbf{diagnosis prediction} task, in which we aim to predict the diagnoses for the patient's potential next visit $V_{T+1}$ given patient's history diagnoses by selecting \vtwo{a set of medical codes} from the medical code ontology $O$, which can be formally described as $f_{DP}:\left\{V_1, V_2, \ldots, V_{T}\right\} \rightarrow V_{T+1}$. The second task is a disease-specific \textbf{heart failure prediction} task, which can be described as a binary classification function $f_{HF}:\left\{V_1, V_2, \ldots, V_{T}\right\} \rightarrow 0,1$. We are more focused and aim to predict whether a patient would encounter heart failure (ICD-9 codes with head \texttt{428}) in any of the future visits. 

\mypar{Input patient record.}
Given an EHR collection of $n$ patients $\{P_1,P_2, \ldots, P_n\}$, patient historical diagnosis can be represented as a sequence of admissions in chronological order $P = \{V_1^{P},V_2^{P}, \ldots, V_T^{P}\}$ where $T$ is the number of existing visits. For a particular visit $V$, the medical judgment made by clinicians as a result of the visit is an \vtwo{unordered set} of diagnoses $V = \{d_1^{V},d_2^{V}, \ldots, d_{|V|}^{V}\}$ in the format of $|V|$ unique medical code ($d \in O$).
\vthree{The task input has two variants, including 1) history diagnosis \textit{codes} only, and 2) additionally providing patient profile (gender, race, medication and family history) as a \textit{natural language sentence}.}
}

\mypar{Medical code ontology as the decision space.}
The International Classification of Diseases (ICD)~\cite{cuadrado2019icd} provides a comprehensive ontology $O$ diseases, symptoms and diagnoses. Each leaf node represents a unique disease/diagnosis and is assigned a unique medical code $c \in \{c_1, c_2, \ldots, c_{|O|}\}$ where $|O|$ is the total number of codes. Diseases are organized into disease groups at multiple levels, represented by non-leaf nodes forming a tree hierarchy $G = \{G_{\text{level}=0}, G_{\text{level}=1}, \dots, G_{\text{level}=depth(O)}\}$.   Assuming the root of $O$ is at level 0, at level $j > 0$, there are $|G_{\text{level}=j}|$ disjoint disease groups, \ie $G_{\text{level}=j} = \{g_1,\ldots,g_{|G_{\text{level}=j}|}\}$.
There is also a one-to-one mapping between a code $c$ and its natural language definition $def_c$.
For example, in version 9 of ICD, the medical code \texttt{250.23} stands for ``Diabetes with hyperosmolarity, type I [juvenile type], uncontrolled''. It belongs to the first-level group for all ``Endocrine, Nutritional, and Metabolic Diseases and Immunity Disorders'', and further belongs to 
the fine-grained disease group ``type I uncontrolled diabetes''. 
\vthree{We use both ICD-9 and ICD-10 coding systems with 13k+ and 68k+ unique codes in this work.}

\spacemagic{\vspace{-1.5em}}
\subsection{Existing Paradigm of Generative LMs}
\label{sec:existing_paradigm}
\spacemagic{\vspace{-0.6em}}

The ordinary formulation of generative LMs takes the input sequence $seq_{in} = t_{1}^{in}, \ldots, t_{|seq_{in}|}^{in}$ and is expected to generate the ground-truth output $seq_{out} = t_{1}^{out}, \ldots, t_{|seq_{out}|}^{out}$.
It produces a probability distribution $P\left(c \mid t_{1:|seq_{in}|}^{in}, \hat{t}_{1:k}^{out}\right)$ over the possible next token ($c \in V$) 
conditioned on both the input sequence and $k$ generated tokens.
Discrete tokens at each autoregressive decoding step are produced by \equationref{eq:decoding}.
The LM is optimized to minimize the cross-entropy loss shown in \equationref{eq:ce_loss} applied on the probability of the \textit{gold} next token conditioned on the \textit{gold} output tokens in the previous segment in a teacher-forcing manner, assuming the $|seq_{out}|$-th token marks the end of the decoding.
\spacemagic{\vspace{-0.3em}}
\begin{equation}\label{eq:decoding}
\hat{t}_{k+1}^{out}=\operatorname{argmax}_{c \in V} P\left(c \mid t_{1:|seq_{in}|}^{in}, \hat{t}_{1:k}^{out}\right)
\spacemagic{\vspace{-0.7em}}
\end{equation}
\spacemagic{\vspace{-0.7em}}
\begin{equation}\label{eq:ce_loss}
\mathcal{L}_{CE} = 
\sum_{k=0}^{|seq_{out}|} -\log P\left(t_{k+1}^{out} \mid t_{1:|seq_{in}|}^{in}, t_{1:k}^{out}\right)
\spacemagic{\vspace{-0.7em}}
\end{equation}

\spacemagic{\vspace{-1.3em}}
\section{\modelname: Learning to Memorize and Rank}
\label{sec:method}

\modelname builds upon a large language model $LM$ after pre-training on a natural language corpus, instruction tuning, and potential alignment process. \modelname is designed to be compatible with numerous generative LM architectures and inherit knowledge obtained through pre-training, including encoder-decoder LM and decoder-only LM. 
There are three steps involved as a pipeline:
1) \vtwo{Fine-tuning} the model to memorize medical codes used to represent the diagnoses; 2) \vtwo{Further optimizing} the model to learn inter-visit causal and temporal relations between patient visits as well as intra-visit patterns from patient history records; 3) During inference, performing autoregressive generation to produce diagnosis predictions given an unseen patient history input.

\spacemagic{\vspace{-0.8em}}
\subsection{Medical Code Memorization}
\spacemagic{\vspace{-0.4em}}
State-of-the-art LLMs struggle to associate medical codes with their correct definitions accurately. 
GPT-4 can only recall 45\% of ICD-9 codes given corresponding definitions (row 3 of \tbref{table:ablation_LM}),
\vtwo{
which may be attributable to the absence of medical codes in the pre-training dataset.
\modelname explicitly teaches $LM$ the semantic information associated with the medical codes and the relationships within the coding system. 
\vthree{
We consider all codes in $O$ as special tokens, each unique medical code has a dedicated token embedding and can be represented by a single token. This design reduces the noise of the learning objectives as the diagnosis probability is equivalent to the token probability.
}
The memorization process \vthree{parameterizes embeddings of the special tokens} and further equips the $LM$ with the necessary external knowledge to facilitate downstream diagnosis prediction.
}
To integrate information about medical codes in $O$ and the natural language knowledge contained in $LM$, we fine-tune $LM$ on synthetic question-answering pairs.

\spacemagic{\vspace{-0.3em}}
\mypar{Bidirectional code and definition memorization.} For each code $c$ and the natural language definition $def_{c}$, we create two input-output pairs. The first pair includes ``What is the definition of ICD-9 code $c$'' as $seq_{in}$ and the target answer ``$def_{c}$'' as $seq_{out}$ to train the model to recall its definition given a code. The second pair helps the model memorize the inverse mapping.
The question-answer pairs are created according to the $O$ ontology being for the downstream task.

\spacemagic{\vspace{-0.3em}}
\mypar{Decision space structure memorization.} 
We further embed code dependencies collectively in $LM$ by training with separate \vtwo{code-category} instances. The curated pairs connect a code to its disease groups at various levels $1, \ldots, depth(O)$ in the code ontology $O$. For example, $seq_{in}$ is ``What is the chapter level disease group of the ICD-9 code 998.51?'', and $seq_{out}$ is ``Injury and Poisoning''.

\begin{figure*}
    \centering
    \includegraphics[width=\textwidth]{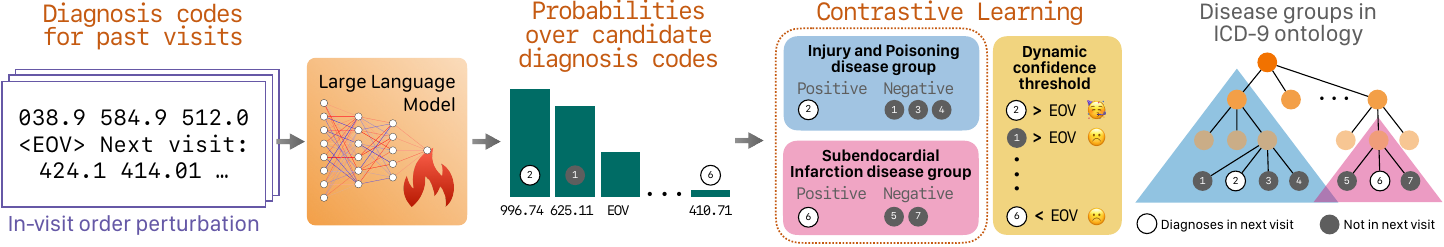}
    \caption{The model design of \modelname. The diagnosis probability distribution is induced from token probabilities. 
    It is optimized with hierarchical contrastive learning and dynamic cross-entropy losses.}
    \label{fig:model}
\end{figure*}

\spacemagic{\vspace{-0.8em}}
\subsection{Seq2seq Data Construction}
\label{sec:seq2seq_data_construction}
\spacemagic{\vspace{-0.6em}}

The second phase aims to equip $LM$ with a temporal and causal understanding of the diagnoses across multiple visits. We train the $LM$ with a collection of sequence-to-sequence training instances $\mathbb{X}=\left\{\mathbf{X}_{1}, \ldots, \mathbf{X}_{n_{\text{patient}}}\right\}$ based on $n_{\text{patient}}$ patient records, where $\mathbf{X}_{i}$ is a set of \vtwo{(diagnosis history, future diagnosis)} pairs created based on patient record $P_i$.
Given the history of a patient containing $T$ visits $P_i = \{V_1^{P_i},\ldots,V_T^{P_i}\}$, we extract $T-1$ pairs of patient history and the expected diagnoses in the next visit to have maximum utilization of the patient records. For each pair, an input sequence is verbalized from 1-to-$k$ visits following $seq_{in} = instruction, vb(V_1^{P_i}), \ldots, vb(V_k^{P_i})$ ($k \in [1, T-1]$). 
\vthree{
Additional patient profile sentences can be inserted following the instructions.
}
A ground-truth output sequence is converted from expected diagnoses in the $(k+1)$-th visit following $seq_{out} = vb(V_{k+1}^{P_i})$.
The verbalizer function $vb$ concatenates the diagnosis codes within each visit to form a token segment for a specific visit and further prepend the starting prompt phrase (``The diagnosis codes for this visit are: '')
and append a special token \texttt{EOV} representing ``the end of the visit''. 

\mypar{Diagnoses order perturbation.} The order of patient visits is crucial to convey the dependent relations as a diagnosis in a later visit is conditioned on the previous diagnoses. However, the order of diagnosis codes \textit{within} a particular visit does not carry cognitive rationale \vtwo{as indicated by EHR dataset documentation and papers~\cite{Johnson2023MIMICIVFreelyAccessible}\reftoapp{ (more details are in \appref{sec:in-visit_code_order})}}.
An ideal model should preserve the \vthree{inter-visit orders while ignoring the intra-visit orders.}
To achieve this goal with a sequential LM, we propose to create $n_{\text{perturb}}$ variants of the input patient history sequences and output diagnosis sequences respectively, leading to $n_{\text{perturb}}^2$ diverse combinations. Each variant keeps the same visit order but randomly shuffles the diagnosis codes within each visit. By observing the data instances with shuffled orders and the same target distribution, we teach the LM to ignore the order of diagnosis codes with a model-agnostic design. 
To summarize, the training sequence-to-sequence data $\mathbb{X}$ contains data instances $\mathbf{X}$ generated according to $n_{\text{patient}}$ patient history records. $\mathbf{X}$ contains $T-1$ groups of data instances with different patient history lengths, each group contains combinations among $n_{\text{perturb}}$ perturbed input sequences and $n_{\text{perturb}}$ perturbed output sequences.

\spacemagic{\vspace{-0.5em}}
\subsection{Learning Inter-visit Reasoning}
\label{sec:intervisit}
\spacemagic{\vspace{-0.4em}}

Up to this point, the created seq2seq data instances can be used to conduct supervised fine-tuning of $LM$ following token-level optimization used in conventional generative LM reiterated in \secref{sec:existing_paradigm}. However, as we analyze theoretically (in \secref{sec:intro}) and demonstrate empirically (line 14/15 of \tbref{table:diagnosis_pred_main}), vanilla generative LM does not handle the diagnosis prediction task well. 
We propose multiple specialized learning objectives to learn the \textit{inter-visit reasoning} to infer upcoming diagnoses and capture \textit{intra-visit diagnosis patterns}. We bridge the sequential modeling capabilities and \vtwo{LM's internal knowledge} with the task property and \vthree{decision} space structure (\eg, ICD hierarchy) for diagnosis prediction.

After encoding $seq_{in}$ containing information on existing hospital visits, the $LM$ starts to generate its prediction of the upcoming visit $\hat{seq_{out}}$. As an immediate step, it produces a probability distribution over the possible next token \vtwo{$t^{out}_1$} conditioned on $seq_{in}$, reflecting the possibility of different tokens in the vocabulary as one of the diagnoses for visit $V_{T+1}$. 
Legit candidate tokens for $t^{out}_1$ are the special code tokens, 
including $\{c_{1}, c_{2}, ..., c_{|O|}\}$. We select the probabilities of all code tokens and then apply softmax, resulting in the probability distribution over the candidate codes
\begin{equation}\label{eq:first_token_prob}
P\left(c \mid t_{1:|seq_{in}|}^{in}\right) = \{p_{c_1}, p_{c_2}, ..., p_{c_{|O|}}\}, c \in O.
\end{equation}
\mypar{Hierarchical contrastive learning.}
We design training objectives to identify the real diagnoses among a group of similar candidate diagnoses. With such a design, the model is forced to understand the subtle differences among neighbor diseases in $O$ and learn to infer upcoming diagnoses from a candidate pool under the same disease group.

For a training instance $\mathbf{X}_{i}$, we first identify all disease groups that the diagnoses of the next visit belong to $G_{\mathbf{X}_{i}} = \{G_{\text{level}=0}, G_{\text{level}=1}, \dots, G_{\text{level}=depth(O)}\}$. Then, for each group $g_k$ at level $j$ ($g_k \in G_{\text{level}=j}$), we identify positive diagnosis codes $g_k^{pos} = \{c_1^{pos}, \ldots, c_{|g_k^{pos}|}^{pos}\}$, which are the diseases in $g_k$ that are diagnosed in the next visit.
We then use all remaining diseases in $g_k$ as negative codes $g_k^{neg} = g_k - g_k^{pos} = \{c_1^{neg}, \ldots, c_{|g_k^{neg}|}^{neg}\}$. Then, we calculate an InfoNCE loss~\cite{oord2018representation,ma-etal-2021-hyperexpan-taxonomy,meng2021coco} term for each group in $G_{\mathbf{X}_{i}}$ and aggregate all the terms to be the aggregated objective $\mathcal{L}_{CL}$.

\begin{equation}\label{eq:infonce_term}
\mathcal{L}_{CL}^{g_{k}} = 
- \log 
\frac{
    \sum_{c_{m}^{pos} \in g_{k}^{pos}} 
    P\left(c_{m}^{pos} \mid t_{1:|seq_{in}|}^{in}\right)
}{
    \sum_{c_{m} \in g_{k}} 
    P\left(c_{m} \mid t_{1:|seq_{in}|}^{in}\right)
}
\end{equation}
\begin{equation}\label{eq:infonce_full}
\mathcal{L}_{CL} = 
\frac{1}{|\mathbb{X}|} 
    \sum_{\mathbf{X}_{i} \in \mathbb{X}} 
    \sum_{G_{\text{level}=j} \in G_{\mathbf{X}_{i}}}
    \sum_{g_{k} \in G_{\text{level}=j}}
    \mathcal{L}_{CL}^{g_{k}}
\end{equation}

The loss term for higher-level groups (where $j$ is smaller) is used to enable the model to recognize disease scopes across a broad spectrum. 
Optimizing the high-level loss mimics the clinician's training process of making differential diagnoses, the ``rough guesses'' of possible diseases.
Loss terms for lower-level groups focus on nuanced comparisons among diseases within the same family, increasing the model's ability to distinguish rare diseases.
The proposed contrastive learning approach is efficient and capable in comparison to in-batch contrastive learning for two reasons: 1) The loss is calculated on the token probability distribution, essential for the typical decoding of generative LM, with no need for additional architecture or forward/backward \vthree{passes}. This ensures efficiency and maximum compatibility with the pre-trained LM. 2) The contradiction for loss calculation pertains to token probabilities, allowing the integration of prediction confidence for each disease into the optimization. This design differs significantly from in-batch contrastive learning, where forward and backward passes must be run for multiple data instances, and batch size significantly limits the size of positive and/or negative samples.

\mypar{Dynamic confidence threshold.} 
To produce a short list of \textit{confident} diagnoses among the full ranking of \textit{all} diagnosis codes, we learn a dynamic confidence threshold to select the most likely predictions.
Existing works apply a fixed threshold to the probability distribution, which is often determined as a hyperparameter observed through the performance of the validation set~\cite{Morid2023TimeSeriesPrediction,Rasmy2021MedBERTPretrainedContextualized}. This widely used strategy makes shortlisting less flexible, and the model tends to play it safe and produces more diagnoses than it should. To model the confidence threshold dynamically, we use a special token \texttt{EOV} to mark the confidence threshold within the token probability ranking list. \texttt{EOV} was appended at the end of the diagnosis sequence of each visit as introduced in \secref{sec:seq2seq_data_construction}. 

The model $LM$ learns the placement of the \texttt{EOV} in two ways. Implicitly, the visit segments in the input sequence demonstrate that the special token \texttt{EOV} represents the end of a visit segment, implying the model should stop generating more diagnosis codes. 
\vtwo{
Training with \texttt{EOV}-ended visit sequence segment, $LM$ naturally learns to assign \texttt{EOV} a higher probability than other code tokens when the model is not confident to make more diagnoses and chooses to generate \texttt{EOV} to end the diagnosis sequence of a particular visit.
}
Explicitly, we design a learning objective to train the $LM$ to place the \texttt{EOV} token at the proper rank of the token probability distribution $P\left(c \mid t_{1:|seq_{in}|}^{in}\right)$. We identify the positive medical codes that do appear in the target visit as \vtwo{$O^{pos}$} and the ones not included as \vtwo{$O^{neg}$ ($O^{pos} + O^{neg} = O$)}. The $\mathcal{L}_{DCE}$ is essentially a dynamic cross-entropy loss that regularizes the probability of each positive code to be not smaller than the probability of \texttt{EOV} and further make sure the probability of each negative code is not larger than $P\left(\text{\texttt{EOV}} \mid t_{1:|seq_{in}|}^{in}\right)$. \vtwo{The optimization of the dynamic confidence threshold applies fine-grained supervision to the probability distribution, enabling effective and efficient diagnosis capability learning with sparse patient data.}
\begin{equation}\label{eq:dce_loss}
\resizebox{\linewidth}{!}{
$
\begin{aligned}
\mathcal{L}_{DCE} = 
\sum_{c \in O^{pos}}
\log 
    \left(
        ReLU(
            P\left(\text{\texttt{EOV}} \mid t_{1:|seq_{in}|}^{in}\right) - 
            P\left(c \mid t_{1:|seq_{in}|}^{in}\right)
        )
    \right)
 \\+
\sum_{c \in O^{neg}}
\log 
    \left(
        ReLU(
            P\left(c \mid t_{1:|seq_{in}|}^{in}\right) - 
            P\left(\text{\texttt{EOV}} \mid t_{1:|seq_{in}|}^{in}\right)
        )
    \right)
\end{aligned}
$
}
\end{equation}

\subsection{Learning Intra-visit Diagnosis Patterns}
\spacemagic{\vspace{-0.4em}}
Besides training the model to reason between visits, there are many implicit rules and latent dependencies buried in the large pool of diagnoses within each visit. For example, within a group of similar diseases, the clinicians normally only choose the most representative code for the patient's status; some diseases might suppress or correlate with other diagnoses. Modeling the intra-visit dependencies enables us to incorporate real-life clinic operation patterns into realistic diagnosis predictions. The prediction made for a specific visit should consider other diagnoses of the same visit. \looseness=-1

To model the intra-visit dependencies, we apply 
the objectives over the token probability distribution introduced in \secref{sec:intervisit} to multiple training instance variants with partial output sequences as conditions. This enables teacher-forcing training. For each ($seq_{in}$, $seq_{out}$) pair in $\mathbf{X}_{i}$ for patient record $P_i$ where the $seq_{out}$ expresses all diagnoses in the visit $V_{k+1}^{P_i}, k \in [1, T-1]$, we create $|V_{k+1}^{P_i}|$ variants to move partial diagnosis results in $seq_{out}$ to be part of the input of $LM$ together with $seq_{in}$. Given the new input including the patient history and $m$ known diagnoses in the upcoming visit, $LM$ produces probability over the candidate medical code $P\left(c \mid t_{1:|seq_{in}|}^{in}, t_{1:m}^{out}\right)$. Since the $m$ known diagnoses have been part of the input sequence, we remove the corresponding medical codes from the positive code set for the calculation of $\mathcal{L}_{DCE}$ and $\mathcal{L}_{CL}$ to prevent the model from generating duplicated codes. Formally, the conditions for probability $P$ in \equationref{eq:first_token_prob}, \equationrefnoprefix{eq:infonce_term}, and \equationrefnoprefix{eq:dce_loss} are \vtwo{$t_{1:|seq_{in}|}^{in}, t_{1:m}^{out}$} instead of \vtwo{$t_{1:|seq_{in}|}$}.
The $m$ known diagnoses in $V_{k+1}^{P_i}$ are removed from $g_k^{pos}$, $O^{pos}$ and added to $g_k^{neg}$ and $O^{neg}$.

\spacemagic{\vspace{-0.8em}}
\subsection{Training and Inference Pipeline}
\spacemagic{\vspace{-0.6em}}
\mypar{Training objectives.}
For code memorization, $LM$ is trained with the ordinary cross-entropy loss in \equationref{eq:ce_loss}. The hierarchical contrastive learning loss (\equationref{eq:infonce_full}) is additionally applied to the instances whose output is a medical code. For the diagnosis prediction task, the $LM$ \vtwo{fine-tuned from the memorization task is further optimized} with the hierarchical contrastive learning loss (\equationref{eq:infonce_full}) and the dynamic cross-entropy loss (\equationref{eq:dce_loss}) on $|V_{k+1}^{P_i}|$ teaching force variants.
Unlike language modeling, no loss has been applied to the reconstruction of the input segment for both fine-tuning stages.
\vtwo{We perform full-parameter fine-tuning.}

\spacemagic{\vspace{-0.3em}}
\mypar{Autoregressive decoding.}
The produced $LM$ can be used for inference on unseen patient history. Given $seq_{in}$, $LM$ performs autoregressive decoding to output discrete diagnosis code with the highest probability in the ranking list for each output step until the \texttt{EOV} token is generated.
\section{Experiments}
\label{sec:experiments}

\subsection{Experimental Setup}
\label{sec:experimental_setup}

\mypar{Datasets.} We use MIMIC-III~\cite{Johnson2016MIMICIIIFreelyAccessible} and MIMIC-IV~\cite{Johnson2023MIMICIVFreelyAccessible} EHR datasets containing patient records to train and evaluate. 
The MIMIC-III dataset focuses on patients eventually admitted to the ICU, while the MIMIC-IV dataset includes both ICU patients and other patients. 
We conduct data preprocessing following previous works~\cite{Lu2022ContextAwareHealthEvent}\vtwo{ and split the train/dev/test sets by patients to avoid information leak}.
\reftoapp{
\vthree{Please refer to \appref{appendix:experimental_setup_details} for data processing details and data statistics.}}

\spacemagic{\vspace{-0.3em}}
\mypar{Metrics.} 
We report the weighted F1 and recall@$k$, where $k$ is the number of top-ranked predictions, and AUC and F1 for diagnosis prediction and heart failure, respectively.

\spacemagic{\vspace{-0.3em}}
\mypar{Baselines.}
\textit{RNN/CNN and attention-based models:}
\textbf{RETAIN} \cite{Choi2016RETAINInterpretablePredictive},
\textbf{Dipole}~\cite{Ma2017DipoleDiagnosisPredictionb},
\textbf{Timeline}~\cite{Bai2018InterpretableRepresentationLearning},
\textbf{HiTANet}~\cite{Luo2020HiTANetHierarchicalTimeAware}, and
\textbf{Deepr}~\cite{Nguyen2017MathttDeeprConvolutional}.
\textit{Graph-based models:}
\textbf{GRAM}~\cite{Choi2017GRAMGraphbasedAttention},
\textbf{G-BERT}~\cite{Shang2019PretrainingGraphAugmented},
\textbf{CGL}~\cite{Lu2021CollaborativeGraphLearning},
\textbf{Chet}~\cite{Lu2022ContextAwareHealthEvent}, and
\textbf{MCDP}~\cite{Li2022MultimodalContrastiveLearning}.
\textbf{KGxDP}~\cite{Yang2023InterpretableDiseasePrediction} formulates each patient as a personalized medical KG, combining medical KGs with patient admission history. Note that additional medical notes are used by CGL, and additional Unified Medical Language System resource~\cite{bodenreider2004unified} is used as external knowledge by KGxDP. 
\reftoapp{
Please refer to \appref{appendix:baseline_details} for details about these baselines.}
\textit{Transformer-based models:} We adapt two encoder-only LM. \textbf{RoBERTa}~\cite{Liu2019RoBERTaRobustlyOptimized} with 125M and \textbf{MedBERT}~\cite{Rasmy2021MedBERTPretrainedContextualized} with 109M parameters and append a $|O|$-way classification head.
We choose MedBERT among other similar encoder-only architectures for medical sequence~\cite{Pang2021CEHRBERTIncorporatingTemporal,Li2023HiBEHRTHierarchicalTransformerBased,Rupp2023ExBEHRTExtendedTransformer} because other models require additional input information such as lab test results which is not available under our setting. \textbf{Seq2seq} uses ordinary generative LM's formulation introduced in \secref{sec:existing_paradigm} to fine-tune a LM to generate diagnosis codes as output. \vtwo{We include definition sentences in the prompt following each code, so these baselines are exposed to the same external knowledge used by \modelname.}

\vthree{
\mypar{Base LMs.}
We use BioMistral~\cite{Labrak2024BioMistralCollectionOpenSource} trained on PubMed Central, 
LLaMA2~\cite{Touvron2023LlamaOpenFoundation}, GPT-2~\cite{RadfordLanguageModelsAre}, T5~\cite{Raffel2023ExploringLimitsTransfer} and Flan-T5~\cite{Chung2022ScalingInstructionFinetunedLanguage} as the base LMs.
}

\begin{table*}[h]
\begin{center}
\resizebox{\linewidth}{!}{
{
\small
\begin{tabular}{rl|rrrrrr|rrrr}
\toprule
\multirow{3}{*}[-4pt]{\#} &
\multirow{3}{*}[-4pt]{\makecell[l]{Model}} & \multicolumn{6}{c|}{Diagnosis Prediction} 
& \multicolumn{4}{c}{Heart Failure}
\\
& & \multicolumn{3}{c|}{MIMIC-III} & \multicolumn{3}{c|}{MIMIC-IV} 
& \multicolumn{2}{c|}{MIMIC-III} & \multicolumn{2}{c}{MIMIC-IV}
\\ \cmidrule{3-12}
& & w-F1 & R@10 & R@20 & w-F1 & R@10 & R@20 & AUC & F1 & AUC & F1
\\ \toprule
\multicolumn{12}{c}{\textbf{\textit{RNN/CNN and attention-based models}}}
\\
1 & Deepr & 18.87 & 24.74 & 33.47 & 24.08 & 26.29 & 33.93 & 81.36 & 69.54 & 88.43 & 61.36\\
2& Dipole & 19.35 & 24.98 & 34.02 & 23.69 & 27.38 & 35.48 & 82.08 & 70.35 & 88.69 & 66.22\\
3& Timeline & 20.46 & 25.75 & 34.83 & \underline{25.26} & \underline{29.00} & \underline{37.13} & 82.34 & 71.03 & 87.53 & 66.07\\
4& RETAIN & 20.69 & \underline{26.13} & 35.08 & 24.71 & 28.02 & 34.46 & \underline{83.21} & 71.32 & \underline{89.02} & 67.38 \\
5& HiTANet & \underline{21.15} & 26.02 & \underline{35.97} & 24.92 & 27.45 & 36.37 & 82.77 & \underline{71.93} & 88.10 & \underline{68.21}\\
\midrule
\multicolumn{12}{c}{\textbf{\textit{Graph-based models}}}
\\
6& G-BERT & 19.88 & 25.86 & 35.31 & 24.49 & 27.16 & 35.86 & 81.50 & 71.18 & 87.26 & 68.04\\
7& GRAM & 21.52 & 26.51 & 35.80 & 23.50 & 27.29 & 36.36 & 83.55 & 71.78 & 89.61 & 68.94\\
8& CGL & 21.92 & 26.64 & 36.72 & 25.41 & 28.52 & 37.15 & 84.19 & 71.77 & 89.05 & 69.36\\
9& MCDP & - & 28.30 & 39.60 & - & 25.80 & 36.10 & - & - & - & -\\
10& Chet & 22.63 & 28.64 & 37.87 & 26.35 & 30.28 & 38.69 & 86.14 & 73.08 & 90.83 & 71.14\\
11& KGxDP & \underline{27.35} & \underline{30.98} & \underline{41.29} & \underline{30.38} & \underline{34.19} & \underline{43.47} & \underline{86.57} & \underline{74.74} & \underline{95.66} & \underline{79.87}\\
\midrule
\multicolumn{12}{c}{\textbf{\textit{Transformer-based models}}}
\\
12& RoBERTa
& 17.39 & 22.84 & 32.07 & 22.54 & 24.89 & 32.38
& 79.74 & 68.28 & 87.03 & 60.21
\\
13& MedBERT
& 19.01 & 23.68 & 34.39 & 24.13 & 25.88 & 33.81
& 81.06 & 69.96 & 88.73 & 61.81
\\
14& Seq2seq (LLaMA2-7B) %
& 18.05 & 18.38 & 23.56 & 20.47 & 20.77 & 24.19
& 77.62 & 66.06 & 85.98 & 59.14
\\
15 & Seq2seq (BioMistral-7B)
& 19.14 & 19.83 & 24.97 
& 22.11 & 22.03 & 26.24
& 78.57 & 67.87 & 87.04 & 61.07
\\
16 & \modelname (LLaMA2-7B)
& 32.77 & 35.94 & 47.48
& 34.64 & 38.16 & 46.94
& 89.49 & 77.21 & 97.26 & 82.31
\\
17 & \modelname (BioMistral-7B)
& \underline{\textbf{33.24}} & \underline{\textbf{36.73}} & \underline{\textbf{49.01}}
& \underline{\textbf{36.16}} & \underline{\textbf{39.57}} & \underline{\textbf{49.09}}
& \underline{\textbf{90.78}} & \underline{\textbf{79.13}} & \underline{\textbf{98.74}} & \underline{\textbf{84.03}}
\\
\bottomrule
\end{tabular}
}
}
\end{center}
\vspace{-0.6em}
\caption{ 
Diagnosis prediction comparison with baselines \vthree{using ICD-9 as the decision space with code-only input} (\%).
}
\label{table:diagnosis_pred_main}
\vspace{-1em}
\end{table*}

\spacemagic{\vspace{-0.6em}}
\subsection{Performance of Diagnosis Prediction}
\spacemagic{\vspace{-0.5em}}

\vthree{
We show the performance comparison on the diagnosis prediction and heart failure prediction tasks (described in \secref{sec:task_formulations}) using ICD-9 as decision space with history diagnosis code as input in \tbref{table:diagnosis_pred_main} and the influence of base pre-trained LM selection in \tbref{table:ablation_LM}. We further show that \modelname can be generalized to richer input with natural language patient profile, and the larger ICD-10 decision space in \tbref{table:diagnosis_pred_nl}.
}

\spacemagic{\vspace{-0.3em}}

\spacemagic{\vspace{-0.3em}}
\mypar{Encoder-only \& vanilla generative LM perform poorly.}
The encoder-only LMs exhibit limited performance (rows 12-13 of \tbref{table:diagnosis_pred_main}), possibly because they do not account for the specialized modeling of intra-visit order and the extensive output space. When employing a vanilla generative LM (rows 14-15), the performance is further diminished. This is attributed to sparse supervision distributed in token-level loss. \vtwo{For each pass, only the probability of the single ground-truth token is optimized following \equationref{eq:ce_loss}, while \modelname optimizes the probabilities of all candidate diagnoses.}

\spacemagic{\vspace{-0.3em}}
\mypar{Gap between zero-shot and fine-tuned LMs.} 
There remains a 20-point deficit in recall@20 comparing the best zero-shot LLM (row 3 of \tbref{table:ablation_LM}) to the fine-tuned model. This underscores the importance of leveraging patient data.\looseness=-1

\spacemagic{\vspace{-0.3em}}
\mypar{\modelname is the state-of-the-art diagnosis prediction model.}
Finally, \modelname achieves significantly better performance in both diagnosis and heart failure prediction tasks on both MIMIC datasets. \modelname exhibits a \vthree{5.89} point higher weighted F1 score and \vthree{almost 8} points higher recall@20 for MIMIC-III compared to the existing best model (row \vthree{17} vs 11 of \tbref{table:diagnosis_pred_main}). In \tbref{table:ablation_LM}, we showcase the diagnosis prediction performance using different pre-trained LMs, noting that even \modelname with GPT-2 large (row 10) achieves comparable performance with the existing best KGxDP.

\begin{table}[h]

\begin{center}
{
\small
\setlength\tabcolsep{4.5pt}
\begin{tabular}{rlH|Hrr|rHr}
\toprule
& & & \multicolumn{3}{c|}{Med. Code Mem.} & \multicolumn{3}{c}{Diagnosis Pred.}
\\
\# & Model & Params \# & Head Code Acc & Code Acc & Def Acc & w-F1 & R@10 & R@20
\\ 
\midrule
\multicolumn{9}{c}{\textbf{\textit{Zero-shot LM}}}
\\
1 & LLaMA2 & 70B
& 16.20 & 4.69 & 0.61
& 5.62 & 6.18 & 15.64
\\
2 & GPT-3.5 & Unknown
& 61.76 & 33.50 & 9.31
& 6.11 & 8.32 & 17.07
\\
3 & GPT-4 & Unknown
& \underline{66.48} & \underline{45.16} & \underline{48.48}
& \underline{6.46} & \underline{10.28} & \underline{21.56}
\\
\midrule
\multicolumn{9}{c}{\textbf{\textit{Fine-tuned encoder-decoder LM}}}
\\
4 & T5 base & 223M
& 90.57 & 81.71 & 1.26
& 20.53 & 21.08 & 30.13
\\
5 & T5 large & 738M
& 92.74 & 85.28 & \underline{2.32}
& 23.19 & 24.96 & 33.85
\\
6 & Flan-T5 base & 248M
& 97.46 & 88.58 & 0.19
& 21.01 & 22.63 & 32.24
\\
7 & Flan-T5 large & 783M
& \underline{97.69} & \underline{89.97} &  0.29
& \underline{25.32} & \underline{25.72} & \underline{35.25}
\\
\midrule
\multicolumn{9}{c}{\textbf{\textit{Fine-tuned decoder-only LM}}}
\\
8& GPT-2 base & 124M
& 0.00 & 0.00 & 95.68
& 23.29 & 24.69 & 32.06
\\
9& GPT-2 medium & 335M
& 0.38 & 0.00 & 98.30
& 25.50 & 26.21 & 34.59
\\
10& GPT-2 large & 774M
& 85.76 & 80.05 & 98.56
& 29.59 & 30.81 & 40.96
\\
11& LLaMA2 7B & 7B
& \underline{\textbf{100.00}} & \underline{\textbf{99.87}} & 99.12 & 32.77 & 35.94 & 47.48
\\
12 & BioMistral 7B &
& \underline{\textbf{100.00}} & 99.61 & \underline{\textbf{99.58}} & \underline{\textbf{33.24}} & \underline{\textbf{36.73}} & \underline{\textbf{49.01}}
\\
\bottomrule
\end{tabular}
}
\end{center}
\vspace{-0.6em}
\caption{ 
Memorization and diagnosis prediction (after fine-tuning on the memorization task) results on MIMIC-III data using different pre-trained LMs.
}
\label{table:ablation_LM}
\vspace{-0.8em}
\end{table}
\begin{table}[h]
\begin{center}
\begin{tabular}{HlH|HHc|HHc}
\toprule
\# & Model & Params \# & w-F1 & R@10 & w NL info & w-F1 & R@10 & w/o NL info
\\ \midrule

1 & Chet &
& & & 17.51
& & & 17.51
\\

2 & Seq2seq (BioMistral 7B) &
& & & 16.31
& & & 13.47
\\
3 & \modelname (BioMistral 7B) & 774M
& & & 43.66
& & & 40.39
\\

\bottomrule
\end{tabular}
\end{center}
\vspace{-0.5em}
\caption{ 
\vthree{Diagnosis prediction results (recall@20, \%) on the MIMIC-IV dataset using ICD-10 as the decision space with or without additional natural language patient profile.}
}
\label{table:diagnosis_pred_nl}
\end{table}

\spacemagic{\vspace{-0.6em}}
\vtwo{
\subsection{Performance on Medical Code Memorization}
}
\spacemagic{\vspace{-0.6em}}
\tbref{table:ablation_LM} shows the evaluation of the memorization results for the ICD-9 medical code system
while using various base LMs. 
\vthree{
We report code and definition accuracy, indicating the proportion of correct output full ICD codes/definitions given their definitions/ICD codes as input by exact match.
} We observed that
\textbf{1) Almost perfect medical code recall using large-enough 7B LM.}
\textbf{2) Pre-trained LLMs alone do not know medical codes well.} GPT models exhibit better memorization of medical codes compared to LLaMA2 (rows 1-3 of \tbref{table:ablation_LM}), \vtwo{but they still lag far behind the fine-tuned models (line 3 vs 12)}. 
\textbf{3) Model scaling-up boosts memorization.} Increasing models' parameters significantly enhances their \vtwo{memorization} capabilities, 
as evidenced by an 80-point improvement in code accuracy from GPT-2 medium to large. However, this does not fully translate into improvement of the same magnitude in diagnosis prediction \vtwo{(row 9 vs 10 in \tbref{table:ablation_LM})}.  %
\textbf{4) Encoder-decoder vs decoder-only.}
Comparing rows 4-7 with rows 8-12 in \tbref{table:ablation_LM}, we observe that encoder-decoder LMs tend to perform well on definition-to-code mapping while performing significantly worse on producing the accurate definition given the code. However, the observation is different for decoder-only LMs who can handle code-to-definition mapping at the early stage.
\vtwo{
\spacemagic{\vspace{-0.3em}}
Derived from these observations, it is optimal to use a large-size decoder-only LM as the backbone for diagnosis prediction.
}

\spacemagic{\vspace{-0.3em}}
\begin{table}[h]
\begin{center}
{
\small
\setlength\tabcolsep{4.5pt}
\begin{tabular}{rl|rHrrr}
\toprule
\# & Method Variant & w-F1 & R@10 & R@20
\\ \midrule
\multicolumn{5}{c}{\textbf{\textit{Knowledge injection approach}}}\\
1 & No external knowledge & \pd{2.33} & \pd{3.76} & \pd{3.54}
\\
2 & Code definition in the prompt
& \pd{1.69} & \pd{1.70} & \pd{2.46}
\\ \midrule
\multicolumn{5}{c}{\textbf{\textit{Training objectives}}}\\
3 & w/o hierarchical contrastive learning
& \pd{10.34} & \pd{8.28} & \pd{10.27} 
\\
4 & -  w/o 0-th level CL loss only
& \pd{9.24} & \pd{7.10} & \pd{8.4} 
\\
5 & -  w/o chapter level CL loss only
& \pd{5.86} & \pd{2.96} & \pd{4.08} 
\\
6 & -  w/o finest level CL loss only
& \pd{7.74} & \pd{5.81} & \pd{6.81} 
\\
7 & w/o dynamic confidence threshold
& \pd{4.10} & \pd{3.07} & \pd{2.57} 
\\ \midrule 
\multicolumn{5}{c}{\textbf{\textit{Outputting strategies}} \modelname = decode (our losses)}
\\
8 & Decode (cross-entropy loss)
& \pd{10.31} & \pd{12.22} & \pd{17.33}
\\
9 & Rank (cross-entropy loss)
& \pd{6.72} & \pd{9.4} & \pd{13.32}
\\
10 & Rank (our losses)
& \pd{2.63} & \pd{3.19} & \pd{3.16}
\\
\bottomrule
\end{tabular}
}
\end{center}
\vspace{-0.5em}
\caption{ 
Ablation study 
on model design choices 
compared with full \modelname (row 16 of \tbref{table:diagnosis_pred_main}) on MIMIC-III dataset.
}
\label{table:ablation}
\vspace{-0.5em}
\end{table}

\spacemagic{\vspace{-0.6em}}
\subsection{Ablation Studies on Method Design}
\spacemagic{\vspace{-0.4em}}

\mypar{Knowledge injection approach.}
In rows 1-2 of \tbref{table:ablation}, 
we observed that simply training the medical code sequence without providing meanings of the codes (row 1) leads to a 3.5-point lower recall@20. Providing the natural language definition of medical code in the input prompt along with the history diagnosis code (row 2 vs 1) is also helpful. However, the NL prompt method suffers from incomplete patient history due to the LM's input length limit, resulting in a 2.5-point lower recall@20 compared to memorization. \vtwo{Fine-tuning for concept memorization is the most effective knowledge injection approach.}

\vtwo{
\spacemagic{\vspace{-0.3em}}
\mypar{Training objectives.}
}%
\vtwo{Results in row 3-7 of \tbref{table:ablation} show that}
removing hierarchical contrastive learning leads to more than a 10-point drop in F1. Among the contrastive terms for disease groups categorized by different granularities, the 0-th level loss (row 4) is the most beneficial, which provides comparisons among the most involved diseases. The finest level loss (row 6) is the second most important, as the chapter-level disease is relatively easier to mine from data, while the fine-grained diagnosis decision involves distinguishing diseases that are similar in manifestation or etiology. \vtwo{Dynamic confidence threshold (row 7) also contributes more than 4-point F1 score improvement.}

\spacemagic{\vspace{-0.3em}}
\mypar{Outputting strategies.}
In rows 8-10 of \tbref{table:ablation}, we explore optimal approaches to produce the diagnosis prediction set.
$LM$ can conduct autoregressive \textit{decoding} to generate diagnosis codes as an output sequence. Alternatively, we can obtain the \textit{ranking} list based on the token probability over the vocabulary of the first output token. \vtwo{Using decoding trained with sparse correct token cross-entropy loss (\secref{sec:existing_paradigm}, row 8) compromises performance by 17 points in recall@20. The confusing in-visit diagnosis code order makes producing the result from the first token ranking list (row 9) a better choice than decoding along. When applying rich supervision with contrastive learning and dynamic confidence threshold, we observe a 10-point higher recall@20 with ranking output (row 10 vs 9). 
The comparison between row 10 and full \modelname validates the effectiveness of intra-visit modeling, yielding a 3-point higher recall@20, where we decode token-by-token conditioned on other diagnoses but with specialized trained token probability for \textit{each} decoding step.}

\spacemagic{\vspace{-0.6em}}
\section{Related Works}
\label{sec:relatedworks}
\spacemagic{\vspace{-0.6em}}

\spacemagic{\vspace{-0.3em}}
\mypar{Diagnosis prediction.}
Existing works leverage structured diagnosis data~\cite{Morid2023TimeSeriesPrediction}. They use sequential models like RNN and LSTM~\cite{Choi2016RETAINInterpretablePredictive,Bai2018InterpretableRepresentationLearning} to model the longitudinal patient history and GNNs to encapsulate spatial features~\cite{Proios2023LeveragingPatientSimilarities,Lu2022ContextAwareHealthEvent}. To inject external knowledge, they conduct multi-task or transfer learning to borrow supervision from other tasks or domains~\cite{Yang2023KerPrintLocalGlobalKnowledge,Zhou2023TransformerbasedRepresentationlearningModel}, use pre-trained embedding to incorporate natural language into initial features~\cite{Wu2023MEGACareKnowledgeguidedMultiview,Bornet2023ComparingNeuralLanguage}, or utilizing external knowledge graphs or ontologies~\cite{An2023KAMPNetMultisourceMedical,Cheong2023AdaptiveIntegrationCategorical,Li2020KnowledgeGuidedDiagnosis}. 
We propose to use the capable LLM architecture to learn patterns from patient history sequences and inject external knowledge with a unified and shared architecture across the pipeline.
Existing works apply contrastive learning on intermediate latent for KG relations~\cite{An2023KAMPNetMultisourceMedical} or patient embedding~\cite{Jeong2023EventBasedContrastiveLearning}, while we apply contrastive learning on diagnosis output space directly.\looseness=-1

\spacemagic{\vspace{-0.3em}}
\mypar{Transformer models for medical event prediction.}
Existing works either handle NL medical notes and other modalities~\cite{Niu2024EHRKnowGenKnowledgeenhancedMultimodal,Zhou2023TransformerbasedRepresentationlearningModel,Wang2023HierarchicalPretrainingMultimodal,Liu2023MultimodalDataMatters}, or they use a non-unified architecture that cannot inherit the pretrained knowledge~\cite{Rupp2023ExBEHRTExtendedTransformer,Li2023HiBEHRTHierarchicalTransformerBased,Pang2021CEHRBERTIncorporatingTemporal,Guo2023EHRFoundationModels} or needs adaptation for downstream tasks~\cite{Steinberg2023MOTORTimeToEventFoundation,Lai2023KEBLMKnowledgeEnhancedBiomedical,Ma2023DICEDataEfficientClinical,Xu2023CanNLIProvidea}. 
\cite{Wang2023DRGLLaMATuningLLaMA,Shoham2023CPLLMClinicalPrediction,Wornow2023EHRSHOTEHRBenchmark} fine-tune the generative LM for classification tasks. We develop a model that is compatible with mainstream LLMs to use the pretrained knowledge and specializes in producing predictions from large diagnosis decision space.

\spacemagic{\vspace{-0.6em}}
\section{Conclusion}
\label{sec:conclusion}
\spacemagic{\vspace{-0.6em}}

\modelname stands out by seamlessly integrating clinical knowledge and addressing the challenges associated with a large candidate space. 
Contrasting learning, tailored to the coding system's hierarchical structure, enables effective distinguishing between accurate and inaccurate diagnosis codes. 
Through validation on MIMIC datasets, \modelname emerges as a leading approach to diagnosis prediction.\looseness=-1

\section*{Acknowledgments}
The work is partially supported by Optum AI, NSF 2200274, 2106859,
2312501, and NIH U54HG012517, U24DK097771.

\bibliography{ma_auto,custom}

\end{document}